\documentclass[10pt,twocolumn,letterpaper]{article}

\usepackage{iccv}
\usepackage{times}
\usepackage{epsfig}
\usepackage{graphicx}
\usepackage{amsmath}
\usepackage{amssymb}
\usepackage{indentfirst}
\usepackage{url}

\usepackage[breaklinks=true,bookmarks=false]{hyperref}

\iccvfinalcopy 

\def\iccvPaperID{****} 

\ificcvfinal\fi
\DeclareRobustCommand*{\IEEEauthorrefmark}[1]{%
    \raisebox{0pt}[0pt][0pt]{\textsuperscript{\footnotesize\ensuremath{#1}}}}
\newcommand{\degree}{^\circ}
\begin{document}

\title{Dual Attention MobDenseNet(DAMDNet) for Robust 3D Face Alignment}
\author{Lei Jiang\textsuperscript{1,3}  \and Xiao-Jun Wu\textsuperscript{1,3,*}\and Josef Kittler\textsuperscript{2}\and
\IEEEauthorrefmark{1}School of IoT Engineering, Jiangnan University 214122,Wuxi, China.\and
\IEEEauthorrefmark{2}Center for Vision, Speech and Signal Processing(CVSSP), University of Surry,GU27XH,Guildford, UK.\and
\IEEEauthorrefmark{3}Jiangsu Provincial Engineering Laboratory of Pattern Recognition and Computational Intelligence,\and Jiangnan University, 214122, Wuxi, China.\\
{\tt\small ljiang\_jnu@outlook.com  xiaojun\_wu\_jnu@163.com j.kittler@surrey.ac.uk}}

\maketitle
\ificcvfinal\thispagestyle{empty}\fi

\begin{abstract}
   3D face alignment of monocular images is a crucial process in the recognition of faces with disguise.3D face reconstruction facilitated by alignment can restore the face structure which is helpful in detcting disguise interference.This paper proposes a dual attention mechanism and an efficient end-to-end 3D face alignment framework.We build a stable network model through Depthwise Separable Convolution, Densely Connected Convolutional and Lightweight Channel Attention Mechanism. In order to enhance the ability of the network model to extract the spatial features of the face region, we adopt Spatial Group-wise Feature enhancement module to improve the representation ability of the network. Different loss functions are applied jointly to constrain the 3D parameters of a 3D Morphable Model (3DMM) and its 3D vertices. We use a variety of data enhancement methods and generate large virtual pose face data sets to solve the data imbalance problem. The experiments on the challenging AFLW,AFLW2000-3D datasets show that our algorithm significantly improves the accuracy of 3D face alignment. Our experiments using the field DFW dataset show that DAMDNet exhibits excellent performance in the 3D alignment and reconstruction of challenging disguised faces.The model parameters and the complexity of the proposed method are also reduced significantly.The code is publicly available at \url{ https://github.com/LeiJiangJNU/DAMDNet}
\end{abstract}

\section{Introduction}

The aim of face alignment is to locate the feature points of the human face, such as the corners of the eyes, the corners of the mouth, tip of the nose. In general it involves fitting a face model to an image and extracting the semantic meaning of facial pixels. This is a fundamental step for many face analysis tasks, such as face recognition \cite{blanz2003face}, face expression analysis \cite{bettadapura2012face} and facial animation \cite{cao2014face,cao2016real}.  In this paper we investigate face alignment in the context of  face disguise detection. The problem of detecting a face disguise is concerned with determining whether a given pair of images belong to the same person even if one of them is subject to a disguise, or to different persons (one of them being an imposter). In view of the importance of this problem, face alignment has been widely studied since the Active Shape Model (ASM) of Cootes in the early 1990s \cite{cootes1994active}. Especially in recent years, face alignment has become a hot topic in computer vision. 

The existing methods of face alignment can be divided into three categories: Constrained Local Model (CLM) methods (e.g., \cite{cootes1994active,saragih2009face}), Active Appearance Model (AAM) methods (e.g., \cite{liu2009discriminative,matthews2004active}) and regression methods (e.g., \cite{cao2014face,valstar2010facial}).

\par 3D face shape reconstruction from 2D image is very challenging by its nature if no prior knowledge is provided.
This is mainly because 2D data does not convey unambiguous depth information.
A common method to solve the problem of monocular 2D face shape reconstruction is to use a set of 3D base shapes to capture the subspace, or a morphological model of face shape variations.
Blanz and Vetter\cite{blanz1999morphable} proposed a comprehensive approach to minimizing the difference between the input 2D image and its 3D face rendering.
Although this method has been able successfully to solve the problem of 3D face reconstruction, it is not friendly to changing lighting conditions, and its computational cost is high.
To overcome this limitation, Blanz \etal.\cite{blanz2004statistical} proposed to predict 3D parameters of a 3D face model from 2D facial feature points by linear regression.
Although this method is efficient, it abandons the most useful information in the image and learns very simple regression functions.
\begin{figure*}
\begin{center}
\includegraphics[width=0.8\linewidth]{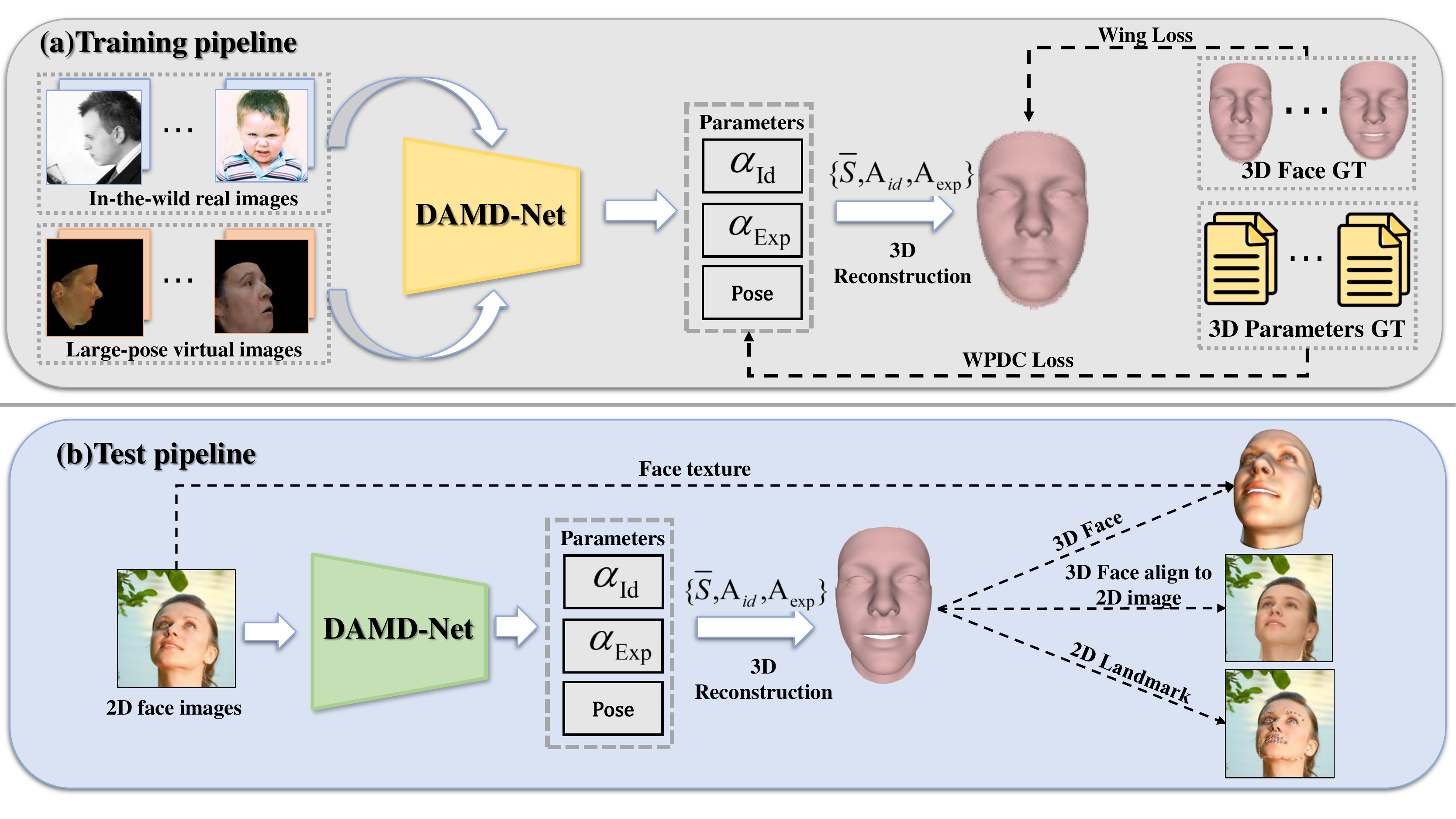} 
\end{center}
   \caption{Overview of our method.As efficient dual attention convolutional neural network(DAMDNet).(a)Training pipeline for a single image 3D face Alignment,(b)Test pipeline. Figure~\ref{fig:details} describes the details of DAMDNet.}
\label{fig:overview}
\end{figure*}
Recently, some innovative methods have been proposed, such as estimating 3D parameters through CNN and related cascaded regression operations, to achieve 3D face reconstruction.
However, the network structure used by these methods is complex and the model parameter space is large, so the network is difficult to train to achieve  convergence.

\par Inspired by the efficiency of MobileNet\cite{howard2017mobilenets}, achieved by the use of the Depthwise Separable Convolution in the network structure, and DenseNet\cite{huang2017densely} strengthened by the transmission of features, we prepose a network structure that extends  both the idea of Depthwise Separable Convolution, and the feature reuse of Densely Connected networks. As dense connection convolution may lead to channel information redundancy, this paper adds a lightweight Channel Attention Mechanism in the network structure, which improves the representation ability of the network without increasing the number of network parameters. 

In convolutional neural networks, in addition to channel feature re-calibration, another important dimension that should be considered is the spatial dimension. For a specific semantic group, it is desirable  and beneficial to identify the  semantic features in the correct spatial location of the original image. Based on the channel attention mechanism, we enhance spatial features by grouping. Spatial Group-wise Enhancement\cite{li2019spatial} is feature re-calibration in spatial dimension. By combining channel and spatial attention mechanisms, the proposed network structure is a dual attention convolutional neural network. 

In order to solve the problem of paucity of training samples in the case of large poses, this paper also presents a side-face data augmentation to enhance the robustness of the network model for arbitrary pose. Extensive experiments are conducted on AFLW dataset\cite{koestinger2011annotated} with a wide range of poses, and the AFLW2000-3D dataset\cite{zhu2016face}, in comparison with a number of  methods. We also provide the means for subjective evaluation by visualizing the 2D/3D face alignment and face reconstruction on the DFW\cite{kushwaha2018disguised,singh2008recognizing} dataset. 

An overview of our method is shown in Figure~\ref{fig:overview}.
\par In summary, our contributions are as follows:
\begin{figure*}
\begin{center}
\includegraphics[width=0.7\linewidth]{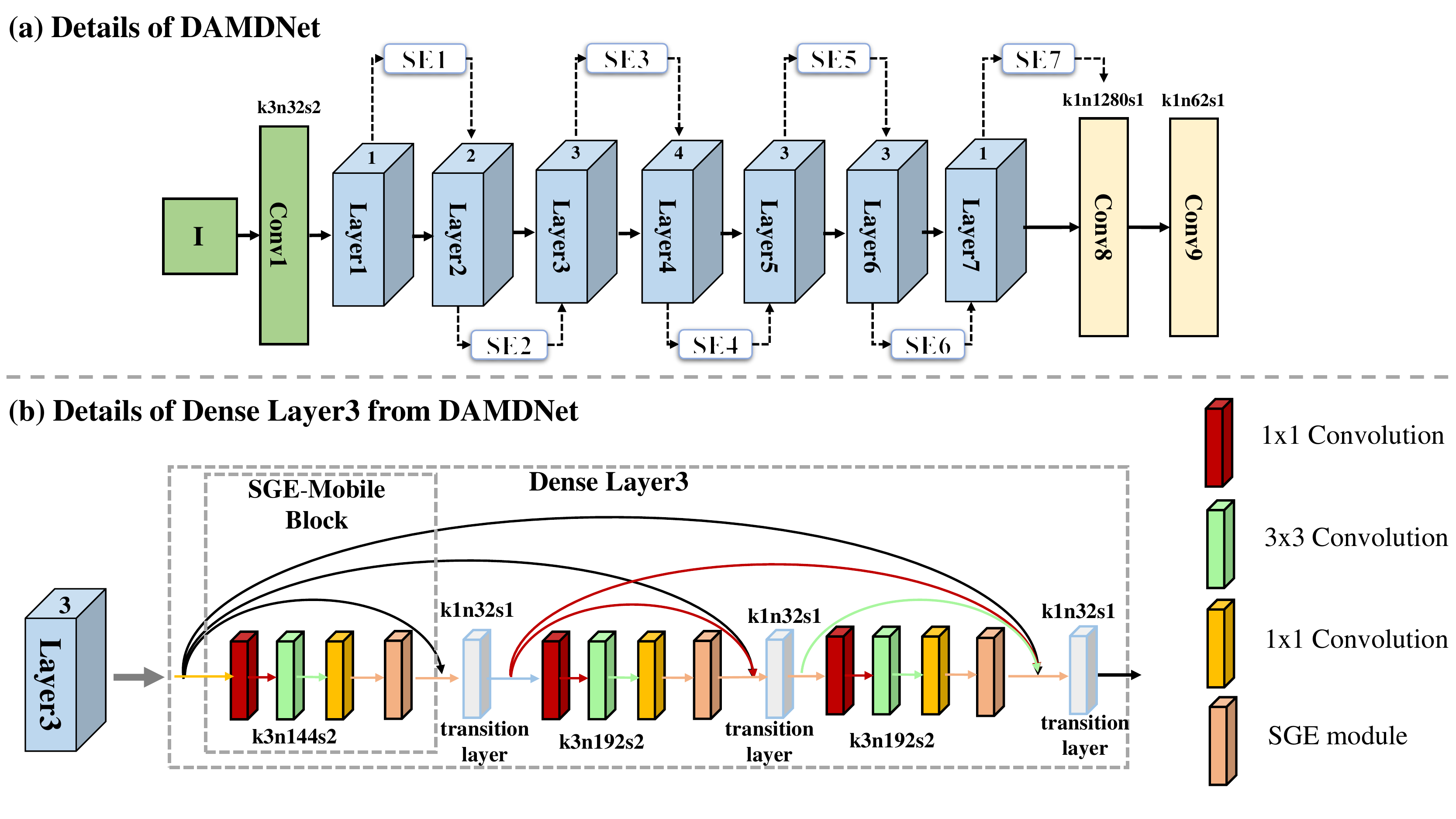} 
\end{center}
   \caption{(a)Details of DAMDNet. k3n64s1 corresponds to the kernel size(k),number of feature maps(n) and stride(s) of conv1. (b)The details of one of the DenseBlock layers, namely $Layer3$. The convolution layer of a set of $1\times1, 3 \times3, 1\times1$ filters and a SGE\cite{li2019spatial} module in DAMDNet as a basic unit called SGE-MobileBlock. The transition layer is the number of channels to match the input and output feature maps.}
\label{fig:details}
\end{figure*}
\par{1)We proposes a novel efficient network structure(DAMDNet).To the best of our knowledge, this is the first time that Depthwise Separable Convolution scheme, a Densely Connected network structure, a Channel Attention Mechanism and  Spatial Group-wise Feature Enhancement are combined to create a DNN novel architecture.} 
\par{2)Different loss functions are used to optimize the parameters of 3D Morphable Model and its 3D vertices. The resulting method can estimate 2D/3D landmarks of faces with an arbitrary pose.}
\par{3)The training data set is augmented by integrating various data enhancement techniques.The face profile technique and virtual sample technique are used to increese its number of the large pose face training data set.}
\par{4)We experimentally demonstrate that our algorithm has significantly improved the 3D alignment performance, compared to the state of the art methods. The  proposed face alignment method can deal with arbitrary poses and it is more efficient.} 
\section{Related Work}
In this section, we  review the prior work in generic face alignment and 3D face alignment.
\subsection{Generic Face Alignment}
Face alignment research can boast many achievements, included the active appearance model(AAM)\cite{cootes2001active,saragih2007nonlinear} and the active shape model(ASM)\cite{cootes2000introduction}.These methods consider face alignment as an optimization problem to find the best shape and appearance parameters, which allow the appearance model to achieve the best possible fit to the input face.The basic idea of the Constrained Local Model (CLM) method~\cite{cristinacce2006feature,asthana2013robust,saragih2011deformable} in the Discriminative approaches category is  to learn a set of local appearance models, one for each landmark.The output of the local models is combined with the help of a global shape model. Cascaded regression gradually refines initial predictions through a series of regressions. Each regression unit relies on the output of the previous regression unit to perform simple image operations. The entire system automatically learns from the training samples\cite{dollar2010cascaded}. The ESR\cite{cao2014face} (Explicit Shape Regression) proposed by Sun \etal. includes three methods, namely two-level boosted regression, shape-indexed features and a correlation-based feature selection method.
\par Besides the traditional models, deep convolutional neural networks have recently been used for feature point localization of faces. Sun \etal.\cite{sun2013deep} were first to use CNN to regress  the raw face image landmark locations,accurately positioning 5 key points of the face from coarse to fine.
The work of \cite{gu20063d} uses the human body pose estimation, and the boundary information for the key point regression. In recent years, most of the landmark detection methods have been adopted some form of  "coarse to fine" strategy. On the other hand, Feng \etal.\cite{feng2018wing} have taken a different approach, using the idea of cascaded convolutional neural networks. A \cite{feng2018wing} compared the commonly used loss functions for face landmark detection, and based on this, the concept of wing loss was proposed.

\subsection{3D Face Alignment}
Although traditional methods provide a guide to successful face alignment, they are affected by non-frontal pose, illumination and occlusion in real-life applications. The most common approach to deal with pose variation is  the multi-view framework \cite{tran2017regressing}, which uses different landmark configurations for different views. For example, TSPM \cite{zhu2012face} and CDM \cite{yu2016face} use the DPM-like \cite{forsyth2014object} method to align faces of different shape models, and finally select the most probable model as the final result. However, since each view requires testing, the computational cost of the multiview approach is always high.
\par Apart from multi-view solutions, 3D face alignment is also a popular approach. 3D face alignment \cite{gu20063d,jourabloo2015pose} aims to fit a 3D morphable model (3DMM) \cite{blanz2003face} to a 2D image. The 3D Morphable Model is a typical statistical 3D face model. It has a clear understanding of 3D faces based on a statistical analysis. Zhu \etal.\cite{zhu2016face} proposed a localization method based on 3D face shape, which solves the problem of some feature points being invisible in extreme poses (such as side faces), as well as the face appearance in different poses varying greatly, making it difficult to locate landmarks. Liu \etal.\cite{jourabloo2016large} used a cascade of 6 convolutional neural networks to solve the problem of locating facial feature points in images of faces with extreme poses by means of  3D face modelling. This method not only predicts the 3D face shape and projection matrix, but also calculates whether each feature point is visible or not.
If a feature point is invisible, the feature block about the invisible point is not used as input, which is difficult to achieve for common 2D face alignment methods. Paper \cite{feng2018joint} designed a UV position map to represent 3D shape features of a complete human face in a 2D. The purpose of 3D face alignment is to reconstruct the 3D face from a 2D image, and then align the 3D face to the 2D image, so that 2D/3D face feature points can be located. Our approach is also based on convolutional neural networks, but we have redesigned the network structure to make it efficient and robust. At the same time, we use different loss functions for 3D parameters and 3D vertices to constrain the semantic information being recovered.
\section{Proposed Method}
In this section we introduce our proposed robust 3D face alignment method, which fits a 3D morphable model using  DAMDNet.
\subsection{3D Morphable Model}
The 3D Morphable model is one of the most successful methods for describing 3D face space. Blanz \etal. \cite{blanz2003face} proposed a 3D morphable model (3DMM) of 3D face based on Principal Component Ananlysis (PCA). It is expressed as follows:
$$S=\overline{S}+A_{id}\alpha_{id}+A_{exp}\alpha_{exp}\eqno(1)$$
where S is a specific 3D face, $\overline{S}$ is the mean face, $A_{id}$ are the principle axes trained on the 3D face scans with neutral expression and $\alpha_{id}$ is the shape parameter vector, $A_{exp}$ are the principle axes trained on the offsets between expression scans and neutral scans and  $\alpha_{exp}$ is the expression parameter vector. The coefficients $\{\alpha_{id},\alpha_{exp}\}$ define a unique 3D face. In this work $A_{id}$ adopted come from the Basel Face Model (BFM)\cite{paysan20093d}  and $A_{exp}$ comes from the FaceWarehouse model\cite{cao2014facewarehouse}.
\par In the process of 3DMM fitting, we use the Weak Perspective Projection to project 3DMM onto the 2D face plane. This process can be expressed as follows:
$$S_{2d}=f*Pr*R*\{S+t_{3d}\}\eqno(2)$$
\par 	where $S_{2d}$ is the 2D coordinate matrix of the 3D face after Weak Perspective Projection, rotation and translation. $f$ is the scaling factor. $Pr$ is the projection matrix $\left(\begin{array}{ccc} 1 & 0 & 0 \\ 0 & 1 & 0 \end{array}\right)$. $R$ is a rotation matrix constructed according to three rotation angles of pitch, yaw and roll respectively. $t_{3d}$  is the 3D translation vector.For the modeling of a specific face, we only need to find the 3D parameters $P=[f,pitch,yaw ,roll,t_{3d},\alpha_{id},\alpha_{exp}]$
\begin{figure}
\begin{center}
\includegraphics[width=1\linewidth]{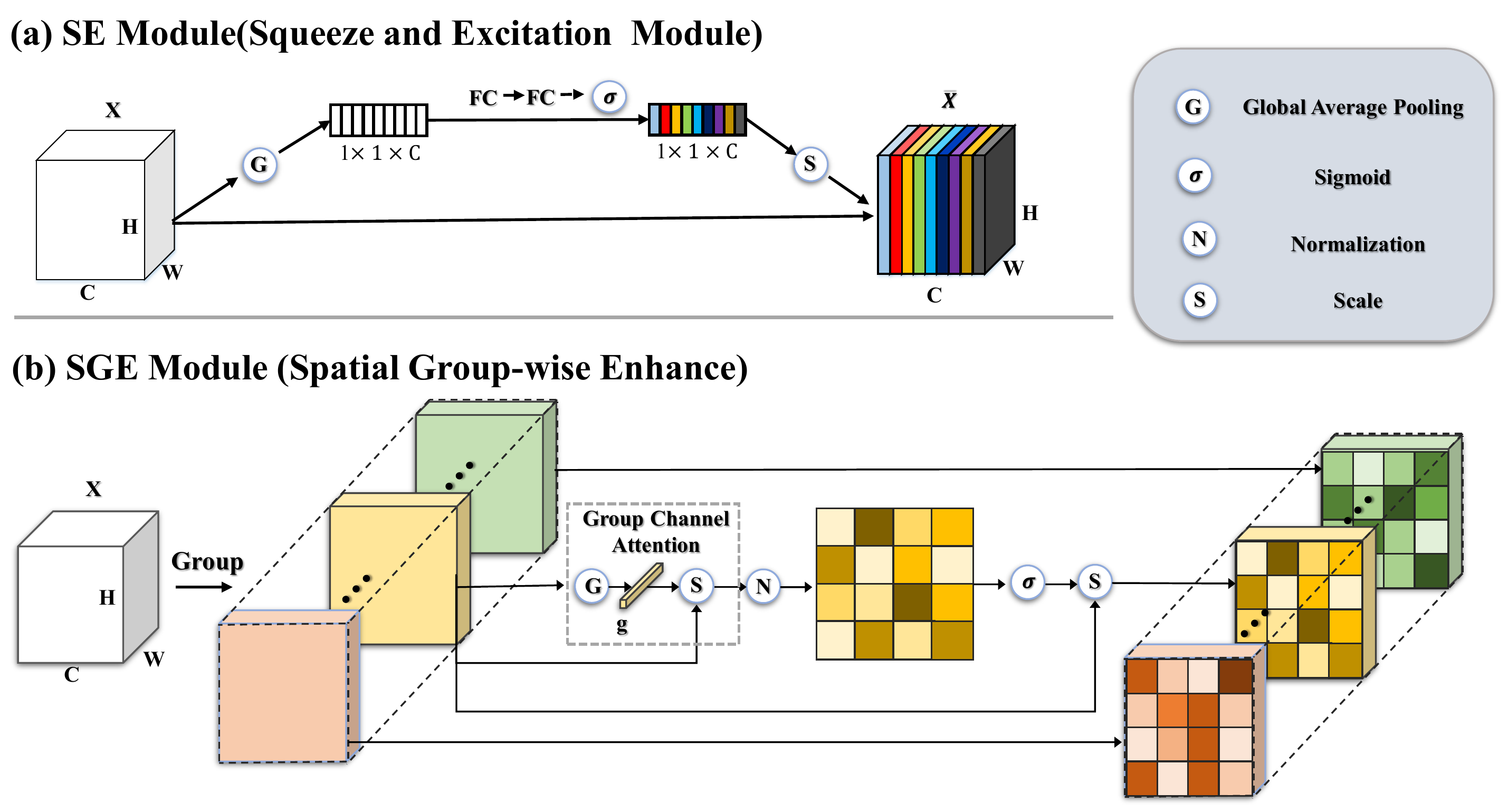} 
\end{center}
   \caption{(a)SE Module(Squeeze and Excitation Module),(b)SGE Module(Spatial Group-wise Enhance).}
\label{fig:att}
\end{figure} 

\subsection{Dual Attention Mechanism}
Extracting the main facial features for the 3D face alignment task is a critical step.  A 2D convolutional neural network typically performs feature extraction in the Channel and Spatial dimensions.This paper enhances the feature representation of convolutional neural networks by adding lightweight attention mechanisms in both, spatial and channel dimensions.
\par In the channel dimension, we opt for  a  SE\cite{hu2018squeeze} attention mechanism module. The SE module uses a new feature recalibration strategy. Specifically, it learns the importance of each feature channel automatically, and enhances the useful features according to the learnt importance measure, and suppresses the non informative features. Figure~\ref{fig:att}(a) describes the basic operation of the SE module.IT first uses a global average pooling layer as a Squeeze operation. Then the two fully connected layers form a Bottleneck structure to model the correlation between the channels and output the same number of weights as the number of input features. A normalized weight between 0 and 1 is obtained by a Sigmoid function to weight each channel.
\par The spatial attention mechanism induces the model to pay more attention to the contribution of the key feature areas of the human face and reduces the influence of other unrelated features. We introduce the SGE(Spatial Group-wise Enhancement)\cite{li2019spatial} mechanism to strengthen the the spatial distribution of facial semantic features. A comprehensive face feature is composed of many sub features, and these sub features are distributed in groups in each feature layer. By generating an attention factor for each group, SGE module can gauge  the importance of each sub feature, and help to suppress noise in a targeted way. This attention factor is determined by the similarity of global and local features within each group, so SGE is very lightweight. Figure~\ref{fig:att}(b) describes the specific computational operations of SGE. First, the features are grouped, and each set of features is spatially compared with the global pooling feature (similarity) to get the initial attention mask.This part we call Group Channel Attention. After, normalizing the attention mask, we obtain the final attention mask through a sigmoid operation, and scale the features of each position to the original feature group.

\subsection{DAMDNet(Dual Attention MobDenseNet) Structure}
The DAMDNet proposed  in this paper applies the depth separable convolution, dense connection, channel attention and spatial attention mechanism to the 3D face alignment task for the first time. The architecture of DAMDNet is illustrated in Figure~\ref{fig:details}(a). Conv1 is a convolution layer with kernel size(k) of 3, stride(s) of 2 and the number of feature maps(n) totalling 32 to extract rough features. $Layer1$ to $Layer7$ are 7 dense blocks for extracting deep features. An SE\cite{hu2018squeeze} module is added between each DenseBlock to explicitly model the interdependencies between feature channels. Figure~\ref{fig:details}(b) shows the details of one of the DenseBlock, $Layer3$. The convolution layer of a set of $1\times1, 3 \times3, 1\times1$ filters and the SGE\cite{li2019spatial} module in DAMDNet form the basic unit called SGE-MobileBlock.
DenseLayer3 contains three sets of SGE-MobileBlock(each SGE-MobileBlock output is cascaded as the input of the next SGE-MobileBlock). As shown in Figure~\ref{fig:details}(b), Layer3 contains three sets of SGE-MobileBlock. In order to match the number of channels connected to the Dense connection, we add a transition layer after each SGE-MobileBlock (the convolution layer filter is $1\times1$), the purpose is to adjust the number of channels in the preview SGE-MobileBlock output feature map.

\subsection{Loss Function}
\par We use two different Loss Functions to jointly train DAMDNet. For predicting 3D parameters we make use of the Weighted Parameter Distance Cost (WPDC) of Zhu \etal. \cite{zhu2016face} to calculate the difference between the ground truth of 3D parameters and the predicted  3D parameters.The basic idea is explicitly to model the importance of each parameter:
$$L_{wpdc}=(P_{gt}-\overline{P})^TW(P_{gt}-\overline{P})\eqno(3)$$
where $\overline{P}$ is an estimate and $P_{gt}$ is the ground truth. The diagonal matrix $W$ contains the weights. For each element of the shape parameter p, its weight is the inverse of the standard deviation that was obtained from the data used in 3DMM training. Our ultimate goal is to accurately obtain 68 landmarks of the human face, For 3D face vertices reconstructed with the estimated 3D parameters, we use Wing Loss\cite{feng2018wing} which is defined as:
$$
\scriptsize
L_{wing}(\Delta{V(P)})=\left\{
\begin{array}{ccl}
\omega\ln(1+|\Delta{V(P)}|/\in) & & {if\;|\Delta{V(P)}| < \omega}\\
|\Delta{V(P)}|-C & &  {otherwise}\\
\end{array}
 \right.\eqno(4)$$where $\Delta{V(P)}=V(P_{gt})-V(\overline{P})$,$V(P_{gt})$ and $V(\overline{P})$ are the ground truth of the 3D facial vertices and the 3D facial vertices reconstructed using the 3D parameters predicted by the network, respectively. $\omega$ and $\in$ are the log function parameters.$C = \omega - \omega\ln(1+\omega/\in)$ is a constant that smoothly links the piecewise-defined linear and nonlinear parts.
\par Overall, the framework is optimized by the following loss function:
$$L_{loss}=\lambda_{1}L_{wpdc}+\lambda_{2}L_{wing}\eqno(5)$$ where $\lambda_{1}$ and $\lambda_{2}$  are parameters, which balance the contribution of $L_{wpdc}$ and $L_{wing}$. The selection of
these parameters will be discussed in the next section.

\subsection{Data Augmentation and Training}
The input to DAMDNet is a 2D image with the facial ROI localized by a face detector. In this paper, we use the Dlib\footnote {http://dlib.net/} SDK for face detection. We first enlarge the detected face bounding box by a factor of 0.25 of its original size and  crop a square image patch of the face ROI, which is scaled to  $120\times120$. DAMDNet outputs a 62-dimensional 3D parameter vector, including 40-dimensional identity parameter vector, 10-dimensional expression parameter vector and 12-dimensional pose vector. We use both real face images and generated face images to train our DAMDNet. We use the same method as \cite{richardson20163d} to generate a virtual face sample with full parameters. The generated face samples contain a large number of large poses.
\par
\begin{figure}
\vspace{-0.6cm} 
\setlength{\abovecaptionskip}{-0.5cm} 
\setlength{\belowcaptionskip}{-0.85cm}
\begin{center}
\includegraphics[width=0.8\linewidth]{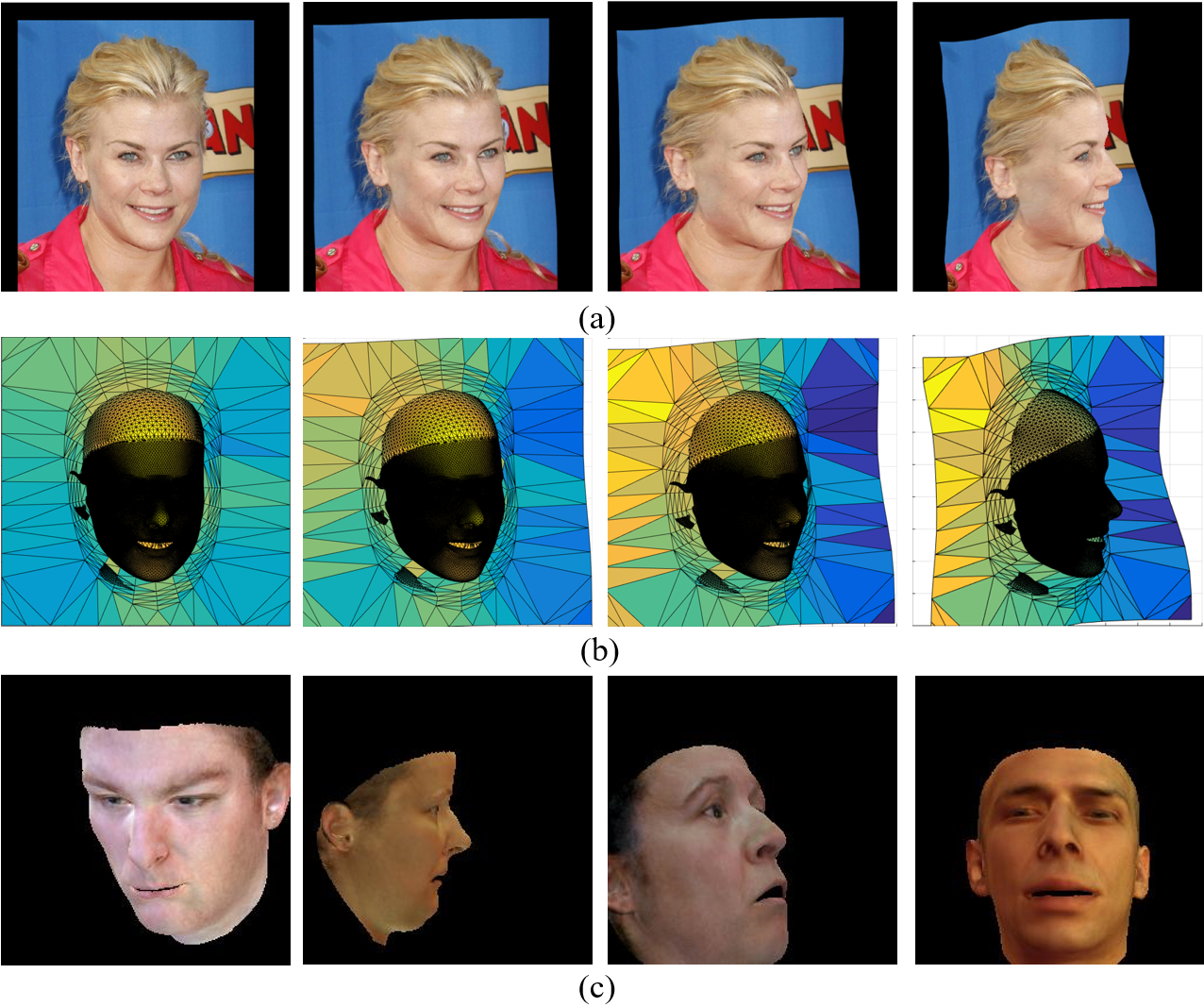} 
\end{center}
   \caption{(a) and (b) from left to right are face images and 3D mesh diagrams rotated by $0\degree$, $15\degree$, $30\degree$, and $60\degree$ around the z-axis;(c)virtual face samples.}
\label{fig:figure5}
\end{figure}

 Most of the current real training data sets contain mages of small and medium poses and unoccluded faces. In order to improve the robustness of the algorithm for arbitrary poses, we conduct a facial profile processing of real face images using the methods proposed by Zhu \etal\cite{zhu2015high}. The face standardization process divides the face image into three regions: the face region, the area around the face, and the background region. Similar to face normalization, the basic idea of face profile is to predict the depth of the face image and generate a contour view that can be rotated in three dimensions. When the depth information is estimated, the face image can be rotated in three dimensions to produce the appearance of larger poses. In this paper, we rotate each real sample by 10 to 90 degrees on the z-axis to generate a face image of new poses.
Figure~\ref{fig:figure5}, (a) and (b) show the effect of a 2D face image rotated by $0\degree$,$15\degree$,$30\degree$ and $60\degree$ respectively, and Figure~\ref{fig:figure5}(c) its 3D mesh
\section{Experiments}
In this section, we evaluate the performance of our method on three common face alignment tasks, face alignment in small and medium poses, face alignment in large poses, and face reconstruction in extreme poses ($\pm 90^o$ yaw angles), respectively.
\subsection{Implementation details}
We use the Pytorch \footnote {https://pytorch.org/} deep learning framework to train the DAMDNet models. The loss weights of our method are empirically set to $\lambda_{1}= 0.5$ and $\lambda_{2}= 1$. In our experiments, we set the parameters of the Wing loss as $\omega = 10$ and $\in = 2$. The Adam solver\cite{kingma2014adam} is employed with the mini-batch size and the initial learning rate set to 128 and 0.01, respectively. There are 680,000 face images in our training set, including 430,000 real face images and 250,000 synthetic face images. Real face images come from AFW\cite{zhu2012face} and LFPW\cite{belhumeur2013localizing} data sets, and various data enhancement algorithms are adopted to expand the datasets. We run the training for a total of 40 epochs. After 15, 25 and 30 epochs, we reduced the learning rate to 0.002, 0.0004 and 0.00008 respectively.
\subsection{Evaluation databases}
We evaluate the performance of our method on three publicly available face data sets AFLW \cite{koestinger2011annotated}, AFLW2000-3D\cite{zhu2016face} and DFW\cite{kushwaha2018disguised,singh2008recognizing}. These AFLW and AFLW2000-3D data sets contain small and medium poses, large poses and extreme poses ($\pm 90^o$ yaw angles). We divide the dataset AFLW and AFLW2000-3D into three sections of $[0^o,30^o], [30^o,60^o], and [60^o,90^o]$  according to the face absolute yaw angle.
\par{\bfseries AFLW}\quad AFLW face database is a large-scale face database including multi-poses and multi-views, and each face is annotated with 21 feature points. This database contains very diverse images, including pictures of various poses, expressions, lighting, and ethnicity. The AFLW face database consists of approximately 250 million hand-labeled face images, of which 59\% are women and 41\% are men. Most of the images are color images, only a few are gray images. We only use the part of extreme pose face images of the AFLW database for qualitative analysis.
\par{\bfseries AFLW2000-3D}\quad AFLW2000-3D is constructed \cite{zhu2016face} to evaluate 3D face alignment on challenging unconstrained images. This database contains the first 2000 images from AFLW and expands its annotations with fitted 3DMM parameters and 68 3D landmarks. We use this database to evaluate the performance of our method for the face alignment task.
\par{\bfseries DFW}\quad Disguised Faces in the Wild (DFW)\cite{kushwaha2018disguised,singh2008recognizing} dataset containing 11,157 images pertaining to 1,000 identities with variations in terms of different disguise accessories. For a given subject there are  four types of images: normal, validation, disguised, and impersonator. We visualized DAMDNet's 3D face alignment effect on the DFW data set, proving that our algorithm also has excellent performance for disguised face.
\subsection{The evaluation metric}
We are given the ground truth 2D landmarks
$U_i$, their visibility $v_i$, and estimated landmarks $\hat{U_i}$ for
$N_t$ test images. Normalized Mean Error (NME) is the average of the normalized estimation error of visible landmarks, defined as,
$$NME=\frac{1}{N_t} \sum_{i}^{N_t}(\frac{1}{d_i|v_i|_1}\sum_{j}^Nv_i(j)||\hat{U_i}(:,j)-U_i(:,j)||)\eqno(6)$$where $d_i$ is the square root of the face bounding box size. Note that normally $d_i$ is the distance of the two centers of the eyes in most prior face alignment work dealing with near-frontal face images.
\begin{table*}[ht]
\renewcommand\arraystretch{1.25}
\centering
\caption{\label{tab:overlapping}The NME(\%) of face alignment results on AFLW and AFLW2000-3D.}
\resizebox{6.8in}{!}{
\begin{tabular}{|c||c|c|c|c|c||c|c|c|c|c|}
\hline & \multicolumn{5}{c||}{AFLW DataSet(21 pts)} & \multicolumn{5}{c|}{AFLW2000-3D DataSet(68 pts)} \\
\hline Method & $[0^o,30^o]$ & $[30^o,60^o]$	& $[60^o,90^o]$	& Mean	&Std	& $[0^o,30^o]$ & $[30^o,60^o]$	& $[60^o,90^o]$	& Mean	&Std	\\
\hline CDM\cite{yu2016face} & 8.150 &	13.020 &	16.170 &	12.440 &	4.040 &	- &	-&	-&	-&	- \\
\hline RCPR\cite{burgos2013robust} & 5.430 &	6.580 &	11.530& 	7.850& 	3.240& 	4.260& 	5.960& 	13.180& 	7.800& 	4.740 \\
\hline ESR\cite{cao2014face} &5.660 &	7.120 &	11.940& 	8.240& 	3.290& 	4.600& 	6.700& 	12.670& 	7.990& 	4.190 \\
\hline SDM\cite{yan2013learn}  &4.750& 	5.550& 	9.340 &	6.550& 	2.450& 	3.670& 	4.940& 	9.760& 	6.120& 	3.210 \\
\hline 3DDFA(CVPR16)\cite{zhu2016face} &5.000& 	\textbf{5.060}& 	6.740& 	5.600& 	0.990& 	3.780& 	4.540& 	7.930& 	5.420& 	2.210 \\
\hline Nonlinear 3DMM(CVPR18)\cite{tran2018nonlinear}& -&	-&	-&	-&	-&	-&	-&	-&	4.700& 	-\\
\hline
\hline Ours & \textbf{4.359}&	5.209& 	\textbf{6.028}&	\textbf{5.199}&	\textbf{0.682}&	\textbf{2.907}	&\textbf{3.830}&\textbf{4.953}&\textbf{3.897}&	\textbf{0.837}\\
\hline
\end{tabular}}
\label{fig:table1}
\end{table*}
\begin{table*}[ht]
\renewcommand\arraystretch{1.25}
\centering
\caption{\label{tab:overlapping}The NME(\%) of face alignment results on AFLW and AFLW2000-3D with the different network structures.}
\resizebox{6.8in}{!}{
\begin{tabular}{|c||c||c||c|c|c|c|c||c|c|c|c|c|}
\hline             &GFLOPs&               Params(M)&                    \multicolumn{5}{c||}{AFLW DataSet(21 pts)} & \multicolumn{5}{c|}{AFLW2000-3D DataSet(68 pts)} \\
\hline Method &           &                                                     & $[0^o,30^o]$ & $[30^o,60^o]$	& $[60^o,90^o]$	& Mean	&Std	& $[0^o,30^o]$ & $[30^o,60^o]$	& $[60^o,90^o]$	& Mean	&Std	\\
\hline RestNeXt50   &1.319&	          23.11&	                 4.599& 	          5.516& 	      6.297& 	         5.471& 	     0.694& 	          3.122& 	        4.065& 	        5.351& 	         4.179& 	0.913  \\ 
\hline MobileNetV2  &\textbf{0.109}&	     \textbf{2.38}&		      4.643& 	          5.581& 	      6.397& 	         5.540& 	     0.716& 	          3.236& 	        4.080& 	        5.181& 	         4.165& 	 \textbf{0.796}  \\  
\hline DenseNet121  &0.800&	          7.02&	                      4.442&              5.249&           6.168&             5.286&            0.705	&               3.051&             3.912&          5.297&           4.087&     0.925  \\
\hline
\hline MDNet        &0.127&	          2.74&		                 4.549&	          5.427&	      6.204&	         5.393&	   \textbf{0.676}&	           3.149&             4.010&          5.270&           4.143&       0.871 \\
\hline AMDNet       &0.128&	          2.75&		                 4.367&	          5.317&	      6.131&	         5.271&	           0.726&	   \textbf{2.879}&             3.906&          4.982&           3.922&      0.858 \\
\hline DAMDNet      &0.125&	          2.76&		         \textbf{4.359}&	  \textbf{5.209}&   \textbf{6.028}&	 \textbf{5.199}&	      0.682&	           2.907	&     \textbf{3.830}& \textbf{4.953}&   \textbf{3.897}&    0.837 \\
\hline
\end{tabular}}
\label{fig:table2}
\end{table*}
\subsection{Comparative evaluation}
\subsubsection{Comparison on AFLW}
\begin{figure}
\begin{center}
\includegraphics[width=0.9\linewidth]{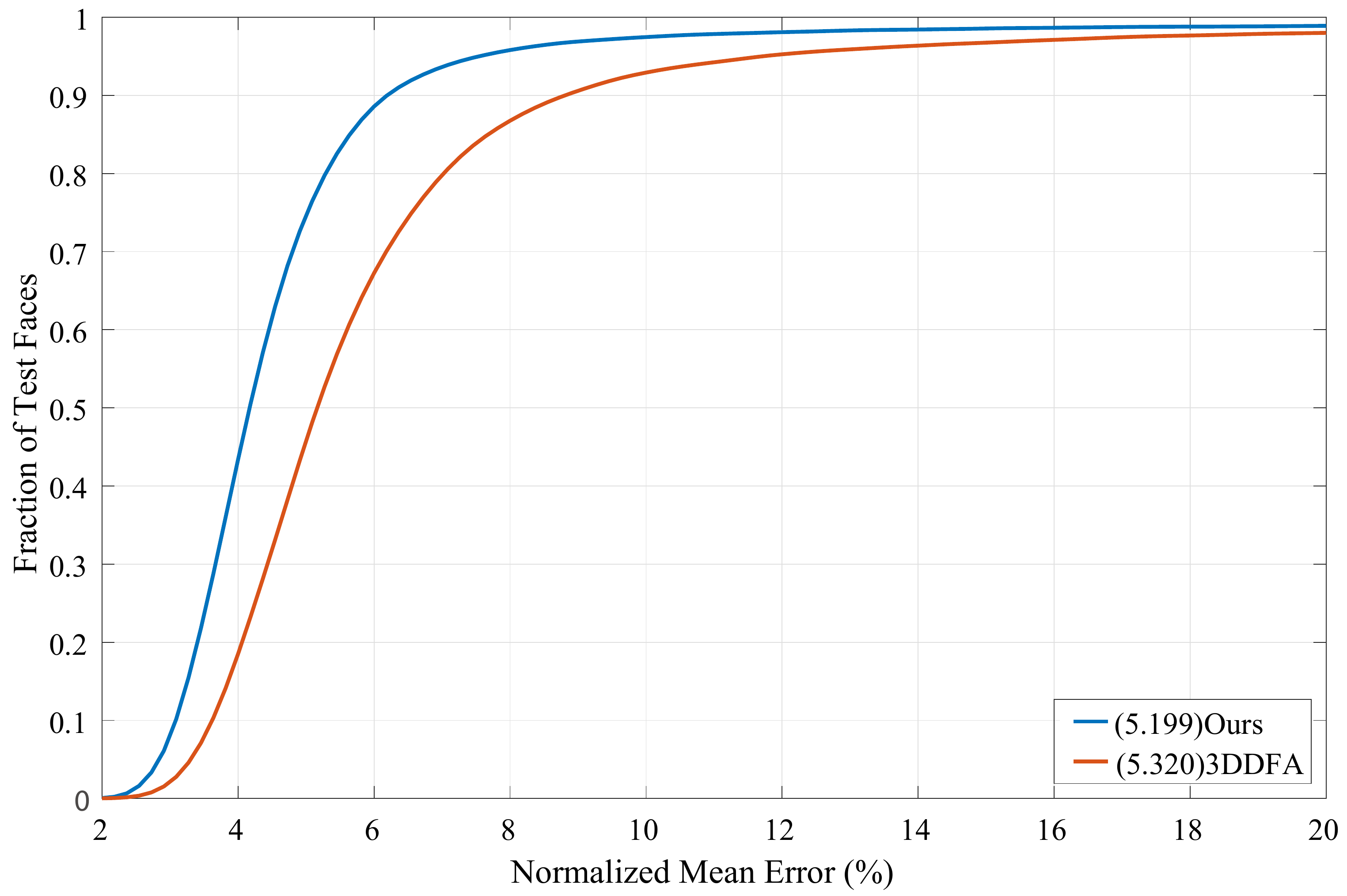} 
\end{center}
   \caption{Cumulative errors distribution (CED) curves on AFLW.}
\label{fig:aflw}
\end{figure}

In the AFLW dataset, 21,080 images were selected as test samples, with 21 landmarks available for each sample. During testing, we divide the test set into 3 subsets according to their absolute yaw angles:  $[0^o,30^o], [30^o,60^o], and [60^o,90^o]$ with 11,596, 5,457 and 4,027 samples respectively. Since a few experiments have been conducted on AFLW, we choose baseline methods for which the code is available, including CDM \cite{yu2016face}, RCPR \cite{burgos2013robust}, ESR \cite{cao2014face}, SDM \cite{yan2013learn},3DDFA\cite{zhu2016face} and nonlinear 3DMM\cite{tran2018nonlinear}.Table~\ref{fig:table1} presents the results, given in terms of NME(\%) of face alignment on AFLW with the best results highlighted. The results of the provided alignment models are identified  with their references. Figure~\ref{fig:aflw} shows the corresponding CED curves. Our CED curve is only compared to the best method in Table~\ref{fig:table1}. Since the best nonliner 3DMM method currently only provides data for the AFLW2000-3D dataset, there is no CED baseline for it. The results show that our algorithm significantly improves the face alignment accuracy in a full range of poses. The minimum standard deviation of our method also proves its robustness to poses changes.

\subsubsection{Comparison on AFLW2000-3D}
\begin{figure}
\begin{center}
\includegraphics[width=0.9\linewidth]{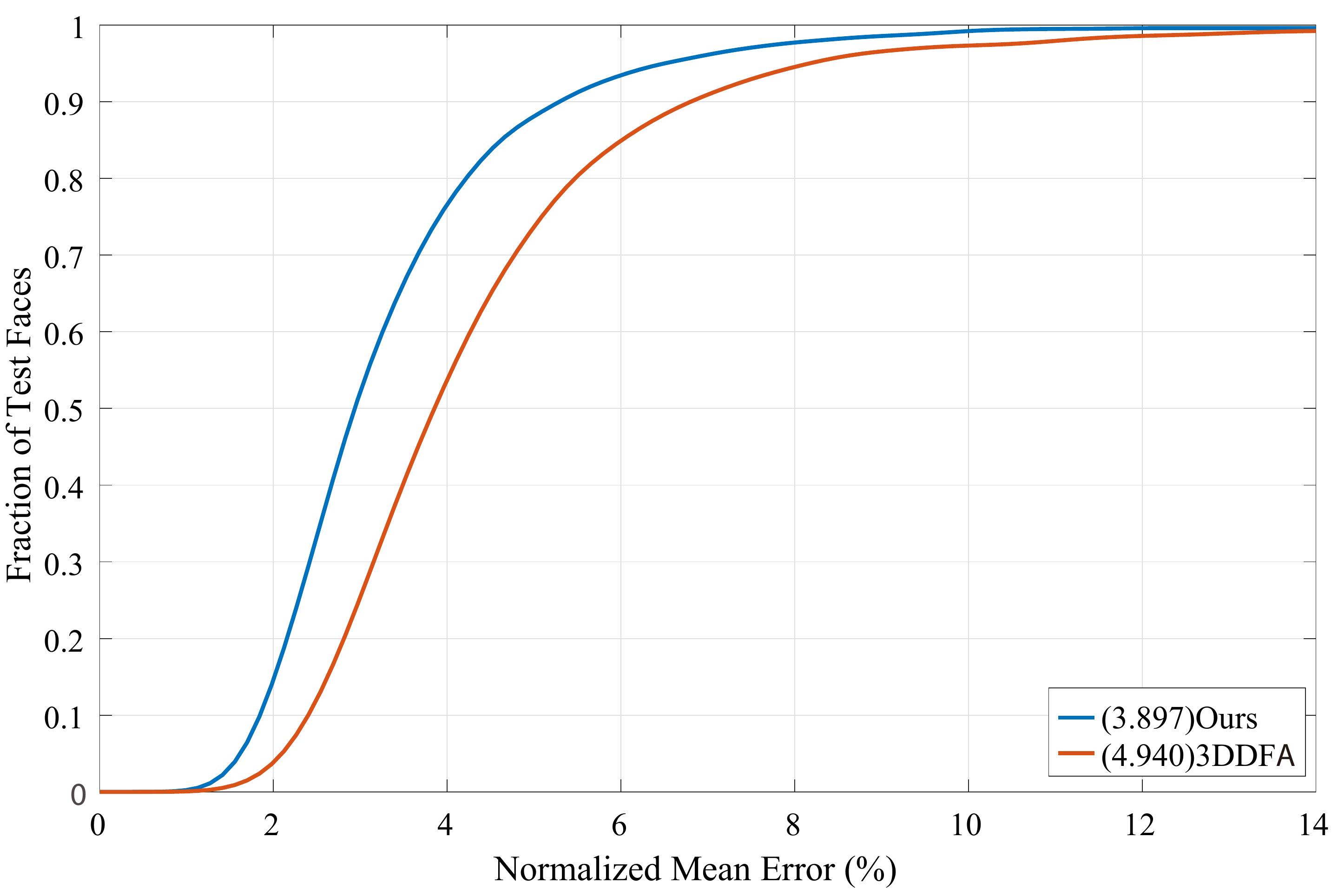} 
\end{center}
   \caption{Comparisons of cumulative errors distribution (CED) curves on AFLW2000-3D.}
\label{fig:aflw2000}
\end{figure}

In the case of  the AFLW2000-3D dataset, 2000 images were selected as test samples. Considering the visible and invisible evaluation, the 3D face alignment evaluation can be transformed to a full landmark evaluation. We divide the test set into 3 subsets according to their absolute yaw angles:  $[0^o,30^o], [30^o,60^o], and [60^o,90^o]$ with 1,312, 383 and 305 samples respectively. Table~\ref{fig:table1} presents the  results (NME(\%)) with the best results highlighted. The results achieved by existing methods are identified by their references. Figure~\ref{fig:aflw2000} shows the corresponding CED curves. Table~\ref{fig:table1} and Figure~\ref{fig:aflw2000} demonstrate that our algorithm also achieves a significant improvement in the prediction of invisible regions, showing good robustness for face alignment in arbitrary poses.
\subsubsection{A visualization experiment performed on DFW}
\begin{figure}
\begin{center}
\includegraphics[width=1\linewidth]{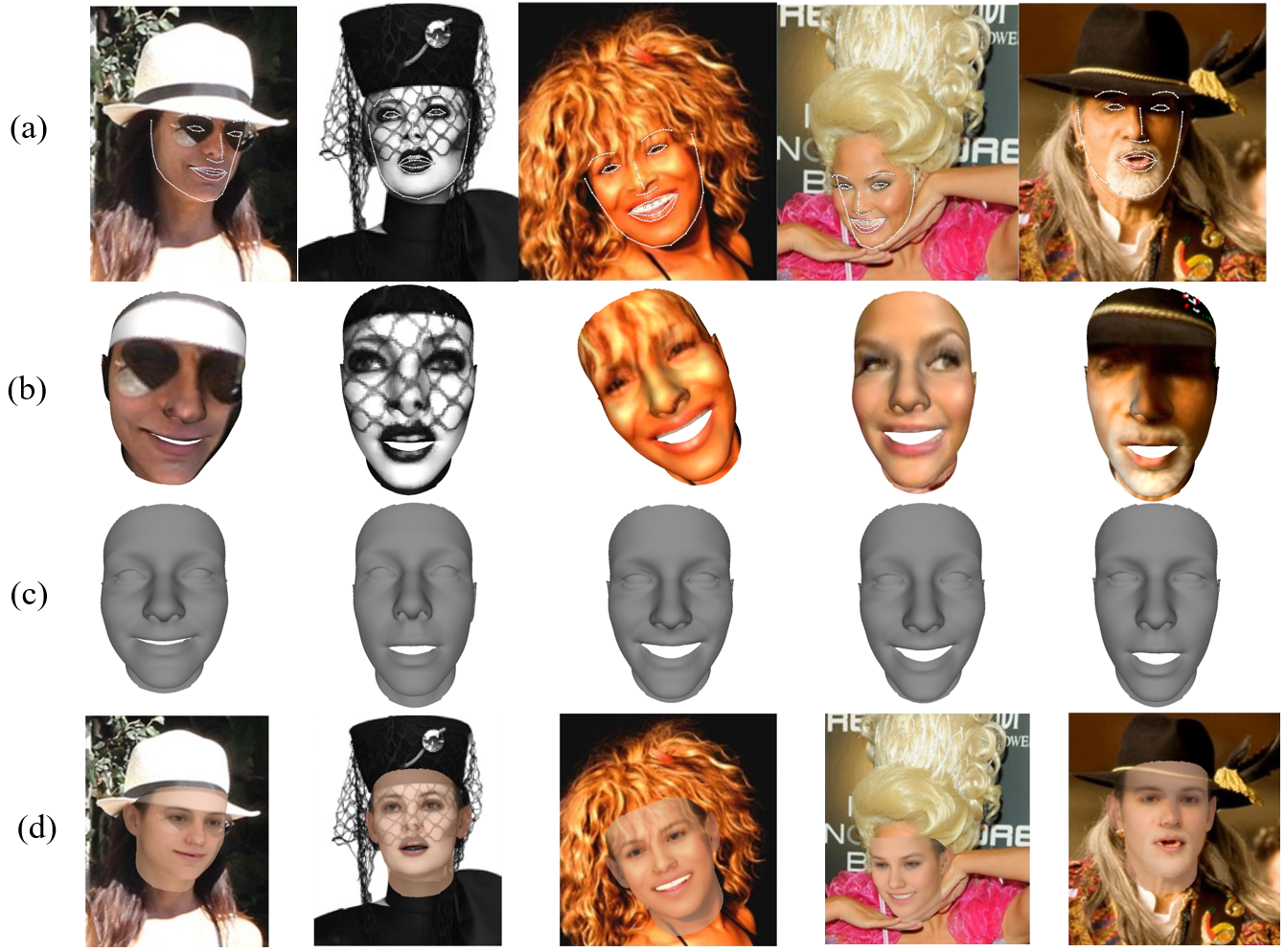} 
\end{center}
   \caption{(a)the landmarks of 2D, (b) the 3D face model with image face texture, (c) the reconstructed 3D face, and (d)  the result of 3d reconstruction of the mean texture of the model using z-buffer projection on the input image.}
\label{fig:dfw}
\end{figure}

In the DFW database, we select some face images for qualitative testing. DFW is by far the most complete data set of disguised faces in the wild. Figure\ref{fig:dfw}  visualizes the results  of our method on DFW, showing (a) 2D landmarks, (b)the fitted 3D face model with image face texture, (c) the reconstructd 3D face and (d)  the result of 3D reconstruction of the mean texture of the model using Z-buffer projection on the input image. Accurate 3D face alignment plays an important role in the next step of disguised face recognition.The results show that our algorithm is robust to disguise. Our algorithm can accurately locate the key points of a face and provide 3D face structure and depth information. This effectively improves the recognition accuracy of the disguised face.

\subsubsection{A comparison of different network structures}
In order to verify the effectiveness of our network structure, we compare our method and the current mainstream neural network structures on the task of face alignment. The experimental network structures include ResNeXt\cite{xie2017aggregated}, MobileNetV2\cite{sandler2018inverted}, DenseNet121\cite{huang2017densely}, and our proposed DAMDNet. To the best of our knowledge, these three popular and efficient network structures are the first applied to the task 3D face alignment. Table~\ref{fig:table2} shows that our DAMDNet achieves a 5\% and 6.7\% reduction in error on the AFLW and AFLW2000 datasets compared to ResNeXt50. In terms of operational efficiency, the GFLOPs complexity is reduced 10.5 times and the number of model parameters is reduced 8.37 times. Compared with DenseNet121, DAMDNet has reduced the error on the AFLW and AFLW2000 data sets by 2.2\% and 4.6\%, the GFLOPs complexity 6.4 times, and the number of model parameters 2.5 times. Compared with MobileNetV2, DAMDNet is higher  both in terms of GFLOPs and the number of network parameters due to the addition of Densely Connected Convolutional and Dual Attention Mechanism in our network structure. However, our model has obvious advantages in terms of accuracy.
\par Similarly, in order to verify the validity of each module of our proposed network structure, we compare MDNet, AMDNet and DAMDNet respectively. Among them, MDNet only combines Depthwise Separable Convolution and Densely Connected structure, AMDNet adds an SE module of Channel Attention Mechanism, and DAMDNet includes Depthwise Separable Convolution, Densely Connected structure and the Dual Attention Mechanism. AMDNet adds a channel attention mechanism based on MDNet, which reduces face alignment error by 2.2\% in the case of the AFLW dataset and 5.1\% for the AFLW2000-3D dataset.
However, the number of GFLOPs and network parameters is increased by 0.78\% and 0.36\% respectively. This result shows that SE module can improve the precision of the network significantly, without adding network parameters and GFLOPs. DAMDNet adds a Spatial Group-wise Feature Enhancement based on AMDNet, which reduces face alignment error by 1.4\% in the AFLW dataset and 0.64\% in the AFLW2000-3D dataset. The GFLOPs is reduced by 2.3\% and the number of parameters is increased by 0.36\%. DAMDNet, which add the SGE module and the Channel Attention Mechanism, significantly improves model efficiency and face alignment accuracy.


\section{Conclusions}
In this paper, we proposed a DAMDNet which solves the problem of 2D/3D face alignment for face images exhibiting a full range of poses. In order to improve the feature expression ability of the model, we incorporated a lightweight attention mechanism for channel and spatial dimensions respectively. We also proposed two novel loss functions to jointly optimize 3D reconstruction parameters and 3D vertices. We use a variety of data augmentation methods and generate a large virtual pose face data set to solve the problem of the large pose training sample imbalance. Our method  achieved the best accuracy on both AFLW, AFLW2000-3D datasets, compared to existing methods. A qualitative evaluation of the proposed method on the DFW dataset showed its promise in dealing with disguised faces.    In comparison to several popular networks, our algorithm achieves a good trade-off between accuracy and efficiency. In the future, will explore the texture and illumination features of face images.

\section*{Acknowledgments}
The paper is supported by the National Natural Science Foundation of China (Grant No.61672265,U1836218),the 111 Project of Ministry of Education of China(Grant No.B12018) and UK EPSRC Grant EP/N007743/1,
Muri/EPSRC/Dstl Grant EP/R018456/1.
{\small
\bibliographystyle{ieee}
\bibliography{ref}
}

\end{document}